\pdfoutput=1

\documentclass[11pt]{article}

\usepackage[final]{coling}

\usepackage{times}
\usepackage{latexsym}

\usepackage[T1]{fontenc}

\usepackage[utf8]{inputenc}

\usepackage{microtype}

\usepackage{inconsolata}

\usepackage{graphicx}
\usepackage[para]{threeparttable}
\usepackage{booktabs}
\usepackage{multirow}
\usepackage{graphicx}
\usepackage{amsmath}
\usepackage{amssymb}
\usepackage{xcolor}
\usepackage{hyperref}
\usepackage{mathtools}
\usepackage[ruled,vlined]{algorithm2e}
\usepackage{tcolorbox}
\usepackage{listings}
\usepackage{enumitem}
\usepackage{makecell}
\definecolor{c1}{HTML}{0049C0}
\usepackage{colortbl}
\usepackage{tabularx}
\definecolor{mygray}{gray}{.95}
\definecolor{mycell}{rgb}{0.85, 0.93, 0.97}
\definecolor{mycelltwo}{RGB}{255, 238, 241}
\usepackage{longtable}
\usepackage{arydshln}
\tcbuselibrary{breakable}
\usepackage{amssymb}
\makeatletter
\renewcommand{\maketag@@@}[1]{\hbox{\m@th\normalsize\normalfont#1}}%
\makeatother

\newcommand{\ie}{\textit{i.e.}}
\newcommand{\eg}{\textit{e.g.}}

\title{\textsc{AdaCQR}: Enhancing Query Reformulation for Conversational Search via Sparse and Dense Retrieval Alignment}

\author{Yilong Lai\thanks{~~Equal Contribution.},\hspace{1.5mm}
   Jialong Wu$^{*}$,\hspace{1.5mm}
   Congzhi Zhang,\hspace{1.5mm}
   Haowen Sun,\hspace{1.5mm}
   Deyu Zhou\thanks{~~Corresponding Author.}\hspace{1.5mm}
   \\
        \hspace{0.5mm}School of Computer Science and Engineering, Key Laboratory of Computer Network\\
        and Information Integration, Ministry of Education, Southeast University, China
    \\
        \texttt{\{yilong.lai, jialongwu, zhangcongzhi, haowensun, d.zhou\}@seu.edu.cn} \\
}

\begin{document}
\maketitle
\begin{abstract}
Conversational Query Reformulation (CQR) has significantly advanced in addressing the challenges of conversational search, particularly those stemming from the latent user intent and the need for historical context. 
Recent works aimed to boost the performance of CQR through alignment. 
However, they are designed for one specific retrieval system, which potentially results in sub-optimal generalization.
To overcome this limitation, we present a novel framework \textsc{AdaCQR}.
By aligning reformulation models with both \textit{term}-based and \textit{semantic}-based retrieval systems, \textsc{AdaCQR} enhances the generalizability of information-seeking queries among diverse retrieval environments through a two-stage training strategy.
Moreover, two effective approaches are proposed to obtain superior labels and diverse input candidates, boosting the efficiency and robustness of the framework.
Experimental results on the TopiOCQA, QReCC and TREC CAsT datasets demonstrate that \textsc{AdaCQR} outperforms the existing methods in a more efficient framework, offering both quantitative and qualitative improvements in conversational query reformulation.\footnote{\raisebox{-2.2pt}{\includegraphics[scale=0.03]{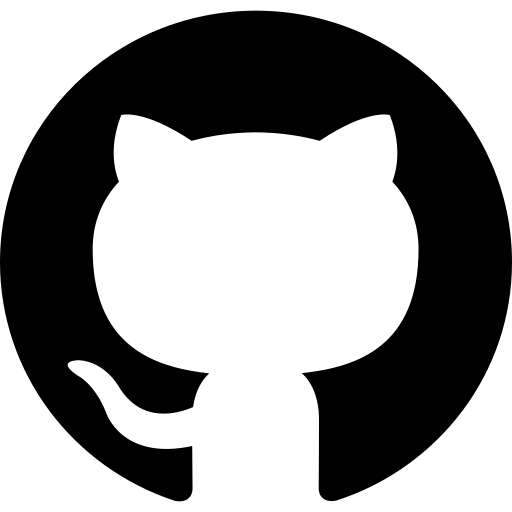}} : \url{https://github.com/init0xyz/AdaCQR}}
\end{abstract}

\section{Introduction}
Conversational search extends the traditional information retrieval paradigms by addressing complex information-seeking requirements through multi-turn interactions ~\citep{conversational_search_2017, dense-retrieval-1, conversational-search-book}. 
A fundamental challenge in conversational search is to discover the latent user intent within the current query and historical context, which complicates the application of off-the-shelf retrievers due to issues such as omissions, ambiguity, and coreference~\citep{qrecc-datasets, topiocqa-datasets}.

\begin{figure}[t]
    \centering
    \includegraphics[width=0.48\textwidth]{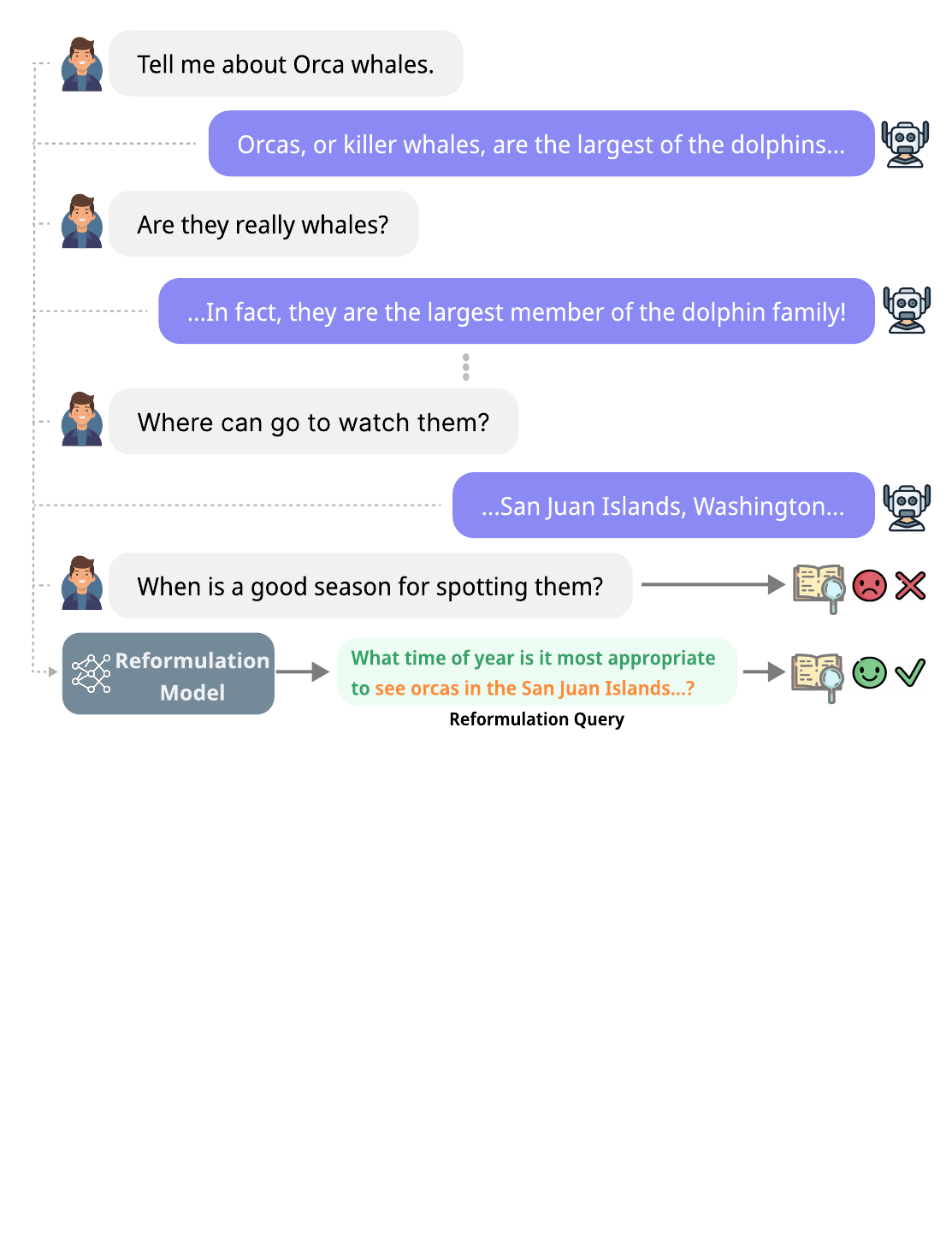}
    \caption{An example of CQR which takes the context and current query as input and generates a decontextualized query as output.} 
    \label{fig:intro}
\end{figure}

Existing methods to address this challenge can be broadly categorized into two types: dense retriever-based and query reformulation-based. For dense retriever-based approaches~\citep{dense-retrieval-1, dense-retrieval-2, dense-retrieval-3, dense-retrieval-4}, long dialogue contexts can be effectively grasped by the dense retriever while incurring retraining costs and lacking the adaptability to sparse retrieval systems like BM25~\citep{bm25}. Query reformulation-based approaches leverage a language model to decontextualize every user's query into a stand-alone query, a process known as conversational query reformulation (CQR), as shown in Figure~\ref{fig:intro}. 
Previous studies have demonstrated the effectiveness of CQR~\cite{ConqRR, ConvGQR, InfoCQR}.

As the training objectives are not aligned with task targets, \ie, minimizing the cross-entropy loss for teacher forcing generation during training while expecting to maximize retrieval metric during inference, subsequent approaches have aimed to enhance the performance of CQR through alignment.
For example, \citet{IterCQR} utilize Minimum Bayes Risk (MBR)~\cite{MBR-2006} based on semantic similarity between the query and gold passage to achieve alignment. \citet{RetPo} create binarized comparisons based on retriever feedback and optimize the reformulation model via direct preference optimization. 
They also tackle the reliance on sub-optimal and costly human-annotated reformulation labels by using Large Language Models (LLMs) to generate labels via iterative prompting or multi-perspective prompting.
However, the methods mentioned above are designed for one specific retrieval system.
For an information-seeking query to be generalized well across both sparse and dense retrieval systems, it must have:
(1) precise \textit{term} overlap (\eg, the presence of key entities in the query) and 
(2) high \textit{semantic} similarity between the document and the query~\cite{sparse-dense-attentional-representation}. 
Both characteristics of the information-seeking query play a crucial role in query reformulation; jointly leveraging them is beneficial.
In addition, previous works leverage reinforcement learning to achieve alignment, but they exhibited stability issues and relied on an explicit reference model ~\cite{ConqRR, IterCQR}.


Therefore, in this paper, we introduce \textsc{AdaCQR}, a novel framework that effectively aligns the training objective with the task target. In specific, \textsc{AdaCQR} aligns the reformulation model and the retrievers from both \textit{term} and \textit{semantic} perspectives to achieve strong generalization abilities in sparse and dense retrieval. 
Furthermore, to address the issues of high complexity and instability inherent in MBR, we employ a two-stage training strategy to achieve alignment, where the reformulation model serves both as a \texttt{generation} model using cross-entropy loss for teacher forcing generation and a \texttt{reference-free evaluation} model using contrastive loss. 


The framework works as follows: 
1) A fusion metric is introduced to evaluate the generalization performance of the reformulated query across various retrieval systems. 
2) By leveraging uncertainty estimation derived from the fusion metric, combined with insights from contrastive learning, we implicitly guide the large language model (LLM) to generate superior reformulation labels, which are then utilized to initialize the reformulation model during the first training stage. 
3) In the second training stage, Diverse Beam Search~\cite{DiverseBeamSearch} is employed to effectively gather multiple candidate reformulation queries. 
The reformulation model is then aligned using a contrastive loss from both term and semantic perspectives, incorporating the candidates and their relative orders based on the fusion metric.

\textsc{AdaCQR} achieves excellent performance
on widely used conversation search datasets, including TopiOCQA ~\cite{topiocqa-datasets}, QReCC ~\cite{qrecc-datasets} and TREC CAsT~\cite{cast19, cast20, cast21}.
Notably, to maintain the smoothness of the overall system, we selected a lightweight model as our backbone, which achieved performance comparable to those approaches fine-tuned on the LLaMA-7B backbone or aggregated multiple candidate queries from proprietary LLM.
Experimental results demonstrate the quantitative and qualitative improvements of our proposed framework.

The contributions of this work are as follows:
\begin{itemize}[itemsep=0.5pt, parsep=0pt]
\vspace{-2mm}
\item We propose \textsc{AdaCQR} to align reformulation models from both \textit{term} and \textit{semantic} perspectives.
\item By leveraging the proposed fusion metric, we can effectively acquire superior labels for generation and collect diverse ordered candidate queries for reference-free evaluation.
\item Extensive experiments on several benchmark datasets conclusively demonstrate our proposed \textsc{AdaCQR} significantly outperforms existing methods, establishing its superiority in performance.
\end{itemize}

\section{Related Work}
\subsection{Conversational Search}

Conversational search improves traditional information retrieval by using iterative, multi-turn interactions to address the complex information needs of users ~\cite{conversational-search-book, mo2024surveyconversationalsearch}. 
A key challenge is understanding the implicit intent of the user, requiring attention to both the current query and its historical context. 
Two main approaches to this problem are conversational dense retrieval (CDR) and conversational query reformulation (CQR).

CDR ~\cite{dense-retrieval-1, CDR_Concept, dense-retrieval-2} aims to improve the representation of the current query along with its historical context by training dense retrievers. 
Recent advancements in CDR have focused on mitigating the influence of irrelevant historical contexts ~\cite{dense-retrieval-3, dense-retrieval-6, mo2024aligningqueryrepresentationrewritten, dense-retrieval-4} and enhancing interpretability ~\cite{dense-retrieval-7, dense-retrieval-8}. 
However, this approach incurs additional training costs and lacks the adaptability to sparse retrieval systems like BM25 ~\cite{bm25}.

Conversely, CQR ~\cite{CQR_concept} concentrates on decontextualizing the query of user into a stand-alone query suitable for use with off-the-shelf retrievers. 
Numerous prior studies have demonstrated the effectiveness of CQR by utilizing human annotations in supervised methods ~\cite{T5QR, CQR_Supervision_3, CQR_Supervision_1} and integrating query expansion models~\cite{ConvGQR, mo-etal-2024-chiq}. 
However, human-annotated labels are costly and reported to be sub-optimal~\cite{dense-retrieval-2, ConqRR}. 
In the era of LLMs, several studies have utilized LLMs to generate query reformulations directly~\cite{InfoCQR, LLM4CS} and obtain reformulation labels for distillation~\cite{IterCQR, RetPo}.
This paper focuses on conversational query reformulation, proposing a novel framework \textsc{AdaCQR} to align with \textit{term}-based and \textit{semantic}-based retrieval systems. 
To overcome the limitations of human annotation, we also developed two effective methods for obtaining superior labels and diverse input candidates.

\begin{figure*}[t]
    \centering
    \includegraphics[width=1.0\textwidth]{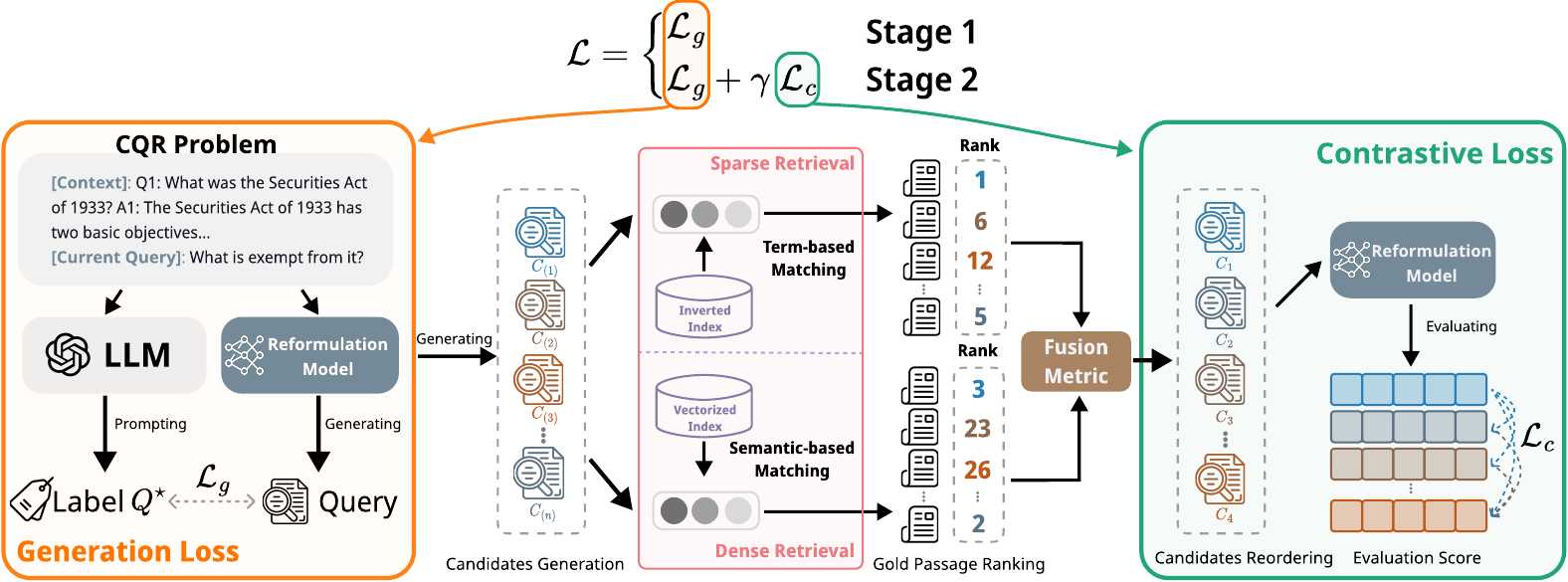}
    \caption{The framework of the proposed \textsc{AdaCQR}. 
    A two-stage training is employed, where Stage 1 involves minimizing generation loss $\mathcal{L}_g$, followed by Stage 2 employing contrastive loss $\mathcal{L}_c$. The evaluation score is a distribution vector defined in Eq.~\eqref{eq:reference-free-evaluation-score}.}
    \label{fig:adacqr-framework}
\end{figure*}

\subsection{Aligning LMs using Feedback}
Aligning language models with feedback involves adjusting their behaviour and outputs based on evaluation feedback~\cite{Alignment-Survey}, employing various reward learning methodologies to provide accurate supervised signals~\cite{PPO, DPO}.

Recent studies have enhanced conversational query reformulation by aligning language models with retriever feedback~\cite{IterCQR, RetPo}. 
\citet{IterCQR} achieve the alignment through minimizing Bayes Risk based on semantic similarity between the query and the gold passage.
\citet{RetPo} leverages LLMs to generate numerous reformulations via multi-perspective prompting, creating binarized comparisons based on retriever feedback and optimizing the reformulation model using DPO ~\cite{DPO}.
However, previous methods struggle with the high cost of generating reformulations with LLMs~\cite{RetPo}, or the instability of MBR~\cite{IterCQR, MBR-Complexity}.
In contrast, our framework utilizes a contrastive loss ~\cite{BRIO} to achieve alignment with retrievers.
To the best of our knowledge, we are the first to employ the language model as a reference-free evaluation model to align retrievers, thereby enhancing stability and reducing complexity.



\section{Method}
\subsection{Task Formulation}
The conversational search task discussed in this paper involves finding the passage most relevant to the intent of the user from a large collection of passages, given the current query of the user and historical context.
To achieve this goal, the CQR task is to utilize a language model $G_\theta$ to condense the current query $q_k$ and historical context $\mathrm{H}_{k-1} = \{q_i,r_i\}_{i=1}^{k-1}$ into a stand-alone query $\hat{Q}_k$, where $q_i$ and $r_i$ denote the query and system answer of the $i$-th turn conversation, with $k$ indicating the current turn.
This decontextualized query $\hat{Q}_k$ is subsequently input into an \textit{off-the-shelf} retrieval system $\mathcal{R}$, which returns a ranked list of the top-$p$ relevant passages. 

For sake of convenience, we formalize the CQR task as a session $\mathcal{P} = \{q, \mathrm{H}\}$, where $q$ represents the user's current query and $\mathrm{H}$ denotes the historical context. The objective is to generate a reformulated query $\hat{Q}$, as discussed in the following sections.

\subsection{Overall Framework}
Figure \ref{fig:adacqr-framework} depicts the overall framework of \textsc{AdaCQR}.
The framework begins to introduce a fusion metric to evaluate the generalization performance of the reformulation queries across sparse and dense retrieval systems(\S\ref{sec:fusion_metric}).
Leveraging the uncertainty estimation based on the fusion metric and the insight from contrastive learning, we select representative examples and implicitly guide LLM to generate superior labels $Q^\star$ to meet the needs of retrievers(\S\ref{sec:CQR_annotation}).
Subsequently, we employ a two-stage training strategy to align the reformulation model with the retrievers. 
In the first stage, we train the reformulation model with a cross-entropy loss $\mathcal{L}_g$ using the superior $Q^\star$ to acquire the basic ability to generate reformulation queries(\S\ref{sec:initialization}).
Afterwards, the model is used to create a diverse set $\mathcal{S}$ comprising candidates queries $C_{(1)}, \cdots, C_{(n)}$ (\S\ref{sec:candidates_generation}), which are then evaluated across sparse and dense retrieval, considering from term and semantic perspectives,  and ranked based on a proposed fusion metric that synthesizes these evaluations.
In the second stage, leveraging the relative order of the candidates, we apply a contrastive loss $\mathcal{L}_c$ (\S\ref{sec:alignment}) to achieve alignment between the reformulation model and the retrievers, where the reformulation model acts as an evaluation model.


\subsection{Fusion Metric for Sparse and Dense Retrieval}\label{sec:fusion_metric}
A good information-seeking query must have precise term overlap and high semantic similarity between the document and the query to generalize well across sparse and dense retrieval~\cite{sparse-dense-attentional-representation}.

In sparse retrieval, the query is tokenized into terms and matches passage based on \textit{term overlap}. 
In contrast, dense retrieval converts the query into an embedding by the encoder and searches passages based on \textit{semantic similarity}.
To measure the generalization ability of the reformulation queries, we input them into sparse and dense retrieval systems and assess their performance based on the ranking of the corresponding gold passages, as illustrated in the central part of Figure~\ref{fig:adacqr-framework}. 



Inspired by reciprocal rank fusion ~\cite{reciprocal_rank_fusion}, we propose a fusion metric to merge the score of sparse retrieval based on term overlap and dense retrieval based on semantic similarity into an optimized score:
\begin{equation}\label{rrf_metric}
\mathrm{M}(\hat{Q},d) = \frac{1}{r_s(\hat{Q}, d)} + \frac{1}{r_d(\hat{Q}, d)}
\end{equation}
where $\hat{Q}$ is a reformulation query, $d$ is the gold passage.
$r_s(q,d)$ and $r_d(q,d)$ represent the rank of the gold passage $d$ within the sparse and dense retrieval results for query $q$, respectively.
The ranking $r_s$ and $r_d$ starts from $1$, indicating the highest-ranked passage.

Leveraging Eq \eqref{rrf_metric}, the generalization performance of the reformulation query $\hat{Q}$ could be evaluated, where a larger $\mathrm{M}(\hat{Q},d)$ indicates better generalization performance for reformulation query $\hat{Q}$ on sparse and dense retrieval systems.

\subsection{Superior Reformulation Annotation}
\label{sec:CQR_annotation}
We leverage LLMs to generate high-quality reformulation labels by conveying the characteristics of effective query reformulation for retrieval.
Given the challenge of explicitly defining optimal reformulation, we propose a prompting strategy that selects representative examples and implicitly guides LLMs to generate reformulation labels based on contrastive learning.

It begins with a \texttt{vanilla model} $G_\pi$ with basic query reformulation ability. 
We employ $G_\pi$ to generate a reformulation candidates set  $\mathcal{S}_\pi$ with diverse beam search for each reformulation session.
Motivated by previous works showing that the annotated most uncertain examples can significantly enhance the effectiveness of in-context learning~\cite{example_selection_1, example_selection_2}, the \textbf{variance}     $\operatorname{Var}(\mathcal{S}_\pi)$, is utilized as a measure for estimating uncertainty in reformulation tasks, where scores for ${S}_\pi$ are computed using the fusion metric described in Eq~\eqref{rrf_metric}.
A higher variance suggests greater instability in the performance of the retriever, indicating a more challenging reformulation problem.
We then selected the top-$m$ representative reformulation problems exhibiting the highest variances on the validation set.

For the representative example annotation, inspired by contrastive learning~\cite{contrastive_prompt_1, contrastive_prompt_2},  we identified the best and worst reformulation candidates from set $\mathcal{S}_\pi$ based on Eq~\eqref{rrf_metric}, denoted as $C_{\text{best}}$ and $C_{\text{worst}}$ to implicitly guide the LLMs in generating labels aligned with the needs of the retrieval system.
We then concatenate the $m$ representative demonstrations, each consists of $(q, \mathrm{H}, C_{\text{best}}, C_{\text{worst}})$, along with task instruction $\mathcal{I}$.
Finally, we employed the LLM to obtain the superior reformulation labels  $Q^\star$ through in-context learning~\cite{GPT3-paper, ICL_survey, xiang-etal-2024-addressing}.
The details of the annotation are presented in Appendix~\ref{appendix:annotation}.

\subsection{Align LMs with Retrievers}\label{sec:AdaCQR_model}


After getting superior reformulation labels using a defined fusion metric, we can align LMs with retrievers through two-stage training.
The reformulation model serves as a standard generation model at the training stage 1. (\S\ref{sec:initialization})
Then we develop a method to generate multiple candidate queries using this trained model. (\S\ref{sec:candidates_generation})
By learning the relative order of these candidates, we implicitly guide the language model to generate queries that meet the requirements of the retrievers. 
Lastly, in training stage 2, the reformulation model serves both as a generation model using cross-entropy loss and a reference-free evaluation model using contrastive loss to achieve alignment. (\S\ref{sec:alignment})
\subsubsection{Training Stage 1 for Initialization} 
\label{sec:initialization}
In the first training stage, we train a language model using the superior reformulation labels to endow it with the basic capability of query reformulation. 
To encourage more diverse generation results, a label smooth cross-entropy loss is used:

\begin{small}
\begin{equation}
\mathcal{L}_1 = \mathcal{L}_g =
\sum_{j=1}^l\sum_{x}p_s(x\mid\mathcal{P}, Q^\star_{<j})\log{p_{G_\theta}(x\mid\mathcal{P}, Q^\star_{<j};\theta)}
\end{equation}
\end{small}
where $\mathcal{P}$ is the reformulation session including current query $q$ and historical context $\mathrm{H}$, $Q^\star_{<j}$ is the first $j$ tokens of the reformulation label $Q^\star$. 
$p_s$ is a label smooth distribution, defined as follows:
\begin{equation}
p_s(x \mid \mathcal{P}, Q_{<j}^\star)= \begin{cases}1-\beta & x=Q_j^\star \\ \frac{\beta}{N-1} & x \neq Q_j^\star\end{cases}
\end{equation}
where $\beta$ is the probability mass parameter, and $N$ is the size of the dictionary. 
Now we have a trained language model $G_\theta$ using cross-entropy loss, which can be used for candidate generation and serves as a reference-free evaluation model during the training at stage 2.

\subsubsection{Candidates Generation for Alignment}\label{sec:candidates_generation}
To efficiently generate a variety of candidates, we utilized Diverse Beam Search ~\cite{DiverseBeamSearch}, an extension of the beam search strategy designed to generate a more diverse set of beam sequences for selection.
Formally, given trained language model $G_\theta$ and reformulation problem $\mathcal{P}$, we generate candidates set $\mathcal{S} = \{C_{(1)}, \cdots, C_{(n)}\}$ with diverse beam search, where $C_{(i)}$ is the candidate of reformulation query, $n$ is the number of candidates.

To align the retrievers from both term-based and semantic-based perspectives with the language model, we define the relative rank order as implicitly supervised signals, utilizing the metric proposed in Eq.\eqref{rrf_metric}, which simultaneously considers both types of retrievers, as follows:
\begin{equation}
C_{(i)} \succ C_{(j)} \iff \mathrm{M}(C_{(i)},d) > \mathrm{M}(C_{(j)},d)
\end{equation}
where $d$ is the gold passage of reformulation problem $\mathcal{P}$.

For reformulation problem $\mathcal{P}$, we now have candidates set $S = \{C_1,\cdots,C_n\}$ and their relative rank order $C_1 \succ C_2 \succ \cdots \succ C_n $, where $C_i$ represents the $i$-th candidate in the sorted order.

\begin{table*}[t!]
    \centering
    \small
    \begin{threeparttable}
    \begin{tabular*}{0.97\textwidth}{clcccc|cccc}
        \toprule
        & &   \multicolumn{4}{c|}{\textbf{TopiOCQA}} &  \multicolumn{4}{c}{\textbf{QReCC}}  \\ 
        \textbf{Type} & \textbf{Query Reform.} & \textbf{MRR} & \textbf{NDCG} & \textbf{R@10}  & \textbf{R@100} & \textbf{MRR} & \textbf{NDCG} & \textbf{R@10}  & \textbf{R@100} \\
        \midrule \midrule
        \multirow{11}{*}{\rotatebox[origin=c]{90}{\textbf{Sparse (BM25)}}} 
        & T5QR (T5-base)  & 11.3 & 9.8 & 22.1 & 44.7 & 33.4 & 30.2 & 53.8 & 86.1 \\
        & CONQRR (T5-base)  & - & - & - & - & 38.3 & - & 60.1 & 88.9 \\
        & EDIRCS (T5-base) & - & - & - & - & 41.2 & - & 62.7 & \underline{90.2}\\
        & IterCQR (T5-base) & 16.5 & 14.9 & 29.3 & 54.1 & 46.7 & 44.1 & 64.4 & 85.5 \\
        & LLM-Aided (ChatGPT) & - & - & - & - & \underline{49.4} & \underline{46.5} & \underline{67.1} & 88.2 \\
        & LLM4CS-REW(ChatGPT) & \underline{16.8} & \underline{15.0} & \underline{31.3} & \underline{56.7} & 35.2 & 32.0 & 55.9 & 84.7 \\
        &  \cellcolor{mygray}{\textsc{AdaCQR} (\textit{Ours}, T5-base)} & \cellcolor{mygray}{\textbf{17.8}} & \cellcolor{mygray}{\textbf{15.8}} & \cellcolor{mygray}{\textbf{34.1}} & \cellcolor{mygray}{\textbf{62.1}} & \cellcolor{mygray}{\textbf{52.4}} & \cellcolor{mygray}{\textbf{49.9}} & \cellcolor{mygray}{\textbf{70.9}} & \cellcolor{mygray}{\textbf{91.0}} \\
        \cmidrule(lr){2-10}
        & ConvGQR (T5-base)$^\heartsuit$ & 12.4 & 10.7 & 23.8 & 45.6 & 44.1 & 41.0 & 64.4 & 88.0 \\
        & \textsc{RetPO} (LLaMA2-7B)$^\heartsuit$  & \textbf{28.3} & \textbf{26.5} & 48.3 & \textbf{73.1} & 50.0 & 47.3 & 69.5 & 89.5 \\
        & LLM4CS-RAR(ChatGPT)$^\heartsuit$ & \underline{27.9} & \underline{26.4} & \underline{48.4} & 71.1 & \underline{51.6} & \underline{49.3} & \underline{75.3} & \underline{92.6} \\
        & \cellcolor{mygray}{\textsc{AdaCQR}\texttt{+Expansion}$^\heartsuit$} & \cellcolor{mygray}{\textbf{28.3}} & \cellcolor{mygray}{\textbf{26.5}} & \cellcolor{mygray}{\textbf{48.9}} & \cellcolor{mygray}{\underline{71.2}} & \cellcolor{mygray}{\textbf{55.1}} & \cellcolor{mygray}{\textbf{52.5}} & \cellcolor{mygray}{\textbf{76.5}} & \cellcolor{mygray}{\textbf{93.7}} \\
        \midrule \midrule
        \multirow{13}{*}{\rotatebox[origin=c]{90}{\textbf{Dense (ANCE)}}}
        & T5QR (T5-base)  & 23.0 & 22.2 & 37.6 & 54.4 & 34.5 & 31.8 & 53.1 & 72.8 \\
        & CONQRR (T5-base)$^\ddagger$ & - & - & - & - & 41.8 & - & 65.1 & \underline{84.7}\\
        & EDIRCS (T5-base) & - & - & - & - & 42.1 & - & 65.6 & \textbf{85.3}\\
        & IterCQR (T5-base) & 26.3 & 25.1 & 42.6 & 62.0 & 42.9 & 40.2 & 65.5 & 84.1 \\
        & LLM-Aided (ChatGPT) & - & - & - & - & \underline{43.5} & \underline{41.3} & \underline{65.6} & 82.3 \\
        & LLM4CS-REW (ChatGPT) & \underline{30.3} & \underline{29.0} & \underline{49.9} & \underline{67.7} & 36.8 & 34.0 & 56.1 & 74.2 \\
        & \cellcolor{mygray}{\textsc{AdaCQR} (\textit{Ours}, T5-base)} & \cellcolor{mygray}{\textbf{32.8}} & \cellcolor{mygray}{\textbf{31.5}} & \cellcolor{mygray}{\textbf{54.6}} & \cellcolor{mygray}{\textbf{73.0}} & \cellcolor{mygray}{\textbf{45.1}} & \cellcolor{mygray}{\textbf{42.4}} & \cellcolor{mygray}{\textbf{66.3}} & \cellcolor{mygray}{83.4} \\
        \cmidrule(lr){2-10} 
        & ConvGQR (T5-base)$^\heartsuit$ & 25.6 & 24.3 & 41.8 & 58.8 & 42.0 & 39.1 & 63.5 & 81.8 \\
        &  \textsc{RetPO} (LLaMA2-7B)$^\heartsuit$  & 30.0 & 28.9 & 49.6 & 68.7 & 44.0 & 41.1 & 66.7 & \textbf{84.6} \\
        & LLM4CS-RAR (ChatGPT)$^\heartsuit$ & \underline{35.4} & \underline{34.4} & \underline{55.2} & \underline{72.2} & \underline{44.7} & \underline{41.8} & \underline{67.2} & \underline{84.0} \\
        & \cellcolor{mygray}{\textsc{AdaCQR}\texttt{+Expansion}$^\heartsuit$} & \cellcolor{mygray}{\textbf{38.5}} & \cellcolor{mygray}{\textbf{37.6}} & \cellcolor{mygray}{\textbf{58.4}} & \cellcolor{mygray}{\textbf{75.0}} & \cellcolor{mygray}{\textbf{45.8}} & \cellcolor{mygray}{\textbf{42.9}} & \cellcolor{mygray}{\textbf{67.3}} & \cellcolor{mygray}{83.8} \\
        \midrule
        \bottomrule 
    \end{tabular*}
    \end{threeparttable}
    \caption{
    Evaluation results of various retrieval system types on the test sets of QReCC and TopiOCQA. 
    The best results among all methods with similar settings are \textbf{bolded,} and the second-best results are \underline{underlined}.
    $\heartsuit$ denotes the method involved in using query expansion.
    $\ddagger$ denotes the baselines utilizing another dual encoder dense retrieval.
    \texttt{+Expansion} denotes the addition of query expansion, details in Appendix \ref{appendix:query_expansion}.
    }
    \vspace{-2mm}
    \label{table:main}
\end{table*}

\subsubsection{Training Stage 2 for Alignment}
\label{sec:alignment}

Now we have sorted candidates  $S$ and trained model $G_\theta$ to perform training at stage 2.
Leveraging the candidates set $\mathcal{S}$ and their relative rank order $C_1 \succ C_2 \succ \cdots \succ C_n$, a contrastive loss ~\cite{BRIO} for alignment:
\begin{equation}
\mathcal{L}_c = \sum_{i=1}^n\sum_{j > i}\max(0, f(C_j)-f(C_i)+(j-i)\times\lambda)
\end{equation}
where $j$ and $i$ are the rank order in the candidates, and $\lambda$ is the margin parameter. 
$f(C)$ represents the length-normalized estimated log-probability, where the language model serves as a reference-free evaluation model:
\begin{equation}\label{eq:reference-free-evaluation-score}
f(C) = \frac{1}{|C|^\alpha}\sum_{t=1}^l\log{p_{G_\theta}}(c_t\mid \mathcal{P}, C_{<t};\theta)
\end{equation}
where $|C|$ and $l$ is the length of candidate, $c_t$ is the generated $t$-th token given reformulation problem and previous $t-1$ tokens, and $\alpha$ is the length penalty parameter.

To ensure the stability of the training process, we employed a multi-task learning loss function, where the language model served as both a generation model and an evaluation model:
\begin{equation}\label{eq:stage2-loss}
\mathcal{L}_2 = \mathcal{L}_g + \gamma \mathcal{L}_c
\end{equation}
where $\gamma$ is the weight of the contrastive loss.

\section{Experiments}
\paragraph{Datasets}
We train and evaluate our model using two widely utilized conversational search datasets: QReCC ~\cite{qrecc-datasets} and TopiOCQA ~\cite{topiocqa-datasets}. 
We also conduct zero-shot experiments on TREC CAsT 19-21~\cite{cast19, cast20,cast21}.
The details of these datasets are shown in Appendix \ref{appendix:datasets}.

\paragraph{Retrieval Systems}
Following prior works in the CQR task ~\cite{ConqRR, ConvGQR, IterCQR, RetPo}, we evaluate \textsc{AdaCQR} using sparse and dense retrieval systems
The sparse retrieval system used is BM25 ~\cite{bm25}. 
For dense retrieval, we use ANCE ~\cite{ANCE}, trained on the MS MARCO ~\cite{MS-MARCO} retrieval task.


\begin{table*}[t]
    \centering
    \small
    \begin{threeparttable}
    \begin{tabular*}{2.05\columnwidth}{lccc|ccc|ccc}
        \toprule
        & \multicolumn{3}{c}{\textbf{CAsT-19}} & \multicolumn{3}{c}{\textbf{CAsT-20}} & \multicolumn{3}{c}{\textbf{CAsT-21}}\\
        \textbf{Query Reform.} & \textbf{MRR} & \textbf{R@10} & \textbf{R@100} & \textbf{MRR} & \textbf{R@10} & \textbf{R@100} & \textbf{MRR} & \textbf{R@10} & \textbf{R@100}\\
        \midrule
        \textsc{T5QR}(T5-base) & 70.1 & - & 33.2 & 42.3 & - & 35.3 & 46.9 & - & 40.8\\
        \textsc{EdiRCS}(T5-base) & 70.9 & - & 35.3 & 43.8 & - & 37.5 & - & - & - \\
        \textsc{LLM4CS-REW}(ChatGPT) & 65.6 & 10.5 & 33.4 & \underline{49.1} & \textbf{16.9} & \textbf{41.3} & \underline{57.7} & \textbf{21.9} & \textbf{55.3}\\
        \textsc{AdaCQR}(\textit{Ours}, T5-base) & \textbf{71.6} & \textbf{12.1} & \textbf{36.7} & \textbf{51.4} & \underline{15.7} & \underline{40.9} & \textbf{58.3} & \underline{21.3} & \underline{54.5} \\
        \midrule
        \textsc{ConvGQR}(T5-base)$^\heartsuit$ & 70.8 & - & 33.6 & 46.5 & - & 36.8 & 43.3 & - & 33.0 \\
        \textsc{LLM4CS-RAR}(ChatGPT)$^\heartsuit$ & 70.6 & 12.6 & 37.7 & \textbf{57.3} & \textbf{19.8} & \textbf{47.6} & \textbf{65.8} & \underline{24.6} & \textbf{59.5}\\
        \textsc{\textsc{AdaCQR}\texttt{+Expansion}}$^\heartsuit$ & \textbf{74.5} & \textbf{13.8} & \textbf{39.2} & \underline{56.6} & \underline{19.2} & \underline{45.6} & \underline{64.2} & \textbf{25.0} & \underline{58.7} \\
        \midrule
        \bottomrule 
    \end{tabular*}
    \end{threeparttable}
    \caption{
    Zero-shot experiment results on TREC CAsT 19-21 datasets.
    The best results among all methods with similar settings are \textbf{bolded,} and the second-best results are \underline{underlined}.
    }
    \vspace{-4mm}
    \label{table:zero-shot}
\end{table*}

\paragraph{Baselines}
To compare with previous CQR baselines, we define two variants of our framework:
\begin{itemize}[itemsep=0pt, parsep=0pt]
    \item \textsc{AdaCQR} generates the reformulation queries through the aligned T5-base model, as illustrated in Section \ref{sec:AdaCQR_model}.
    \item \textsc{AdaCQR}\texttt{+Expansion} concats the reformulation queries generated by \textsc{AdaCQR} and query expansions generated by vanilla LLaMA2-7B leveraging the pseudo answers and keywords expansion techniques as described in the Appendix \ref{appendix:query_expansion}.
\end{itemize}
We categorize previous works into two groups: \textit{query reformulation} and \textit{query reformulation with expansion$^\heartsuit$}.
In the query reformulation section, we compare \textsc{AdaCQR} with the fine-tuned T5-base models including: T5QR~\cite{T5QR}, CONQRR~\cite{ConqRR}, EDIRCS~\cite{EDIRCS}, IterCQR~\cite{IterCQR} and LLM-based prompting methods including LLM-Aided~\cite{InfoCQR} and LLM4CS~\cite{LLM4CS} under the Rewriting Prompting(REW) setting.
In the query reformulation with expansion section, we compare \textsc{AdaCQR}\texttt{+Expansion} with T5-based model ConvGQR\footnote{
\textsc{AdaCQR} (T5-base) without expansion, is showing superior performance than ConvGQR$^\heartsuit$ whose expansions are generated by another T5-base model.}~\cite{ConvGQR}, the fine-tuned LLaMA2-7B model \textsc{RetPO}~\cite{RetPo} and LLM-based prompting method LLM4CS~\cite{LLM4CS} with the Rewrite-and-Resonse(RAR) setting.

The details regarding \textbf{baselines}, \textbf{implementation} and \textbf{evaluation metrics} are provided in Appendix~\ref{appendix:baselines}, Appendix~\ref{appendix:implementation_details} and Appendix~\ref{appendix:eval}, respectively.


\subsection{Main Results}
To evaluate the efficacy of our framework, we conducted comprehensive experiments datasets with the \textsc{AdaCQR} and baselines, presented in Table~\ref{table:main}.
We consider three kinds of backbones as baselines: the T5-based, the LLaMA2-7B-based, and the ChatGPT-based.

The results under the without expansion setting demonstrate that \textsc{AdaCQR} significantly outperforms previous models utilizing T5-base as the backbone. 
Among methods with expansion, the exceptional performance of \textsc{RetPO} and LLM4CS-RAR, which use LLaMA2-7B and ChatGPT as the backbone, respectively, can be attributed to the inherently strong common-sense reasoning capabilities of backbone models.
\textsc{AdaCQR} with expansion achieves results comparable to \textsc{RetPO} and LLM4CS-RAR, with only a slight disadvantage in the R@100 metric, while empirically outperforming them in other settings.
Specifically, \textsc{AdaCQR}\texttt{+Expansion} shows superior performance in dense retrieval on the TopiOCQA, attaining the best MRR (38.5), NDCG (37.6), R@10 (58.4), and R@100 (75.0).
The reported results show significant improvements with the $t$-test at $p < 0.05 $ overall compared to baselines in all the settings on QReCC and TopiOCQA (except \textsc{AdaCQR}\texttt{+Expansion} with ANCE on TopiOCQA). 

These results underscore the efficacy and generalizability of \textsc{AdaCQR} in enhancing retrieval performance across different retrieval systems.

\subsection{Zero-shot Results}
To access the generalization performance of \textsc{AdaCQR}, we conduct the experiment on TREC CAsT datasets under a zero-shot setting, shown in Table~\ref{table:zero-shot}.
Among all non-expansion methods, \textsc{AdaCQR} achieves the best performance across all metrics on the CAsT-19 and the highest MRR scores on the CAsT-20 and CAsT-21.
Among methods with expansion, \textsc{AdaCQR} still performs best on the CAsT-19 dataset and demonstrates comparable results to the LLM4CS-RAR approach on the CAsT-20 and CAsT-21.
The strongest competitor in expansion setting, LLM4CS-RAR, leverages advanced proprietary LLM and candidates queries aggregation to enhance its expansion capability, so it is reasonable that \textsc{AdaCQR} falls slightly short. 
Achieving comparable performance further demonstrates the strength of our method. 
These results provide solid evidence of the generalization capabilities of our approach.

\subsection{Ablation Study}
\begin{table}[t]
    \centering
    \small
    \begin{threeparttable}
    \begin{tabular*}{0.94\columnwidth}{clccc}
        \toprule
        \textbf{Type} & \textbf{Abaltion Variants} & \textbf{MRR} & \textbf{R@10}\\
        \midrule
        \midrule 
        \multirow{5}{*}{\rotatebox[origin=c]{90}{\textbf{Sparse}}} 
        & \textsc{AdaCQR} (\textit{Ours}) & \textbf{52.4} & \textbf{70.9}\\
        \cmidrule(l){2-4}
        & \quad \cellcolor{mycell}{w/o. Contrastive Loss} & \cellcolor{mycell}{43.3} & \cellcolor{mycell}{62.8}\\
        & \quad \cellcolor{mycelltwo}{w/o. Fusion Metric} &  \cellcolor{mycelltwo}{50.5} &  \cellcolor{mycelltwo}{67.7} \\
        & \quad  \cellcolor{mycelltwo}{w/o. Sparse Rank} &  \cellcolor{mycelltwo}{50.9} &  \cellcolor{mycelltwo}{69.7} \\
        & \quad \cellcolor{mycelltwo}{w/o. Dense Rank} &  \cellcolor{mycelltwo}{51.6} &  \cellcolor{mycelltwo}{70.5} \\
        \midrule
        \multirow{5}{*}{\rotatebox[origin=c]{90}{\textbf{Dense}}} 
        & \textsc{AdaCQR} (\textit{Ours}) & \textbf{45.1} & \textbf{66.3}\\
        \cmidrule(l){2-4}
        & \quad \cellcolor{mycell}{w/o. Contrastive Loss} & \cellcolor{mycell}{38.5} & \cellcolor{mycell}{58.9}\\
        & \quad  \cellcolor{mycelltwo}{w/o. Fusion Metric} &  \cellcolor{mycelltwo}{42.4} &  \cellcolor{mycelltwo}{63.7} \\
        & \quad  \cellcolor{mycelltwo}{w/o. Sparse Rank} &  \cellcolor{mycelltwo}{43.5} &  \cellcolor{mycelltwo}{64.2} \\
        & \quad  \cellcolor{mycelltwo}{w/o. Dense Rank} &  \cellcolor{mycelltwo}{42.9} &  \cellcolor{mycelltwo}{63.0} \\
        \midrule
        \bottomrule 
    \end{tabular*}
    \end{threeparttable}
    \caption{
    Ablation study for alignment and ranking of \textsc{AdaCQR} on QReCC dataset.
    }
    \label{table:ablation}
\end{table}

\begin{table}[t]
    \centering
    \small
    \begin{threeparttable}
    \begin{tabular*}{0.83\columnwidth}{clccc}
        \toprule
        \textbf{Type} & \textbf{Query Reform.} & \textbf{MRR} & \textbf{R@10}\\
        \midrule
        \midrule 
        \multirow{5}{*}{\rotatebox[origin=c]{90}{\textbf{Sparse}}}  &
        \texttt{Human Rewrite} & 39.8 & 62.7 \\
        & \texttt{LLM Rewrite} $Q^\star$  & 45.4 & 65.5\\
        \cmidrule(l){2-4}
        & \textsc{AdaCQR} \\
        & \quad Trained on \#\texttt{HR} & 44.9 & 63.7 \\
        & \quad Trained on $Q^\star$ & \textbf{52.4} & \textbf{70.9}\\
        \midrule
        \multirow{5}{*}{\rotatebox[origin=c]{90}{\textbf{Dense}}}  & \texttt{Human Rewrite} & 38.4 & 58.6 \\
        & \texttt{LLM Rewrite} $Q^\star$ & 40.1 & 60.2\\
        \cmidrule(l){2-4}
        & \textsc{AdaCQR}\\
        & \quad Trained on \#\texttt{HR} & 41.0 & 60.5 \\
        & \quad Trained on $Q^\star$ & \textbf{45.1} & \textbf{66.3} \\
        \midrule
        \bottomrule 
    \end{tabular*}
    \end{threeparttable}
    \caption{
    Ablation study for reformulation labels on QReCC dataset.
    The superior LLM reformulation $Q^\star$ is obtained by in-context learning (detailed in \S \ref{sec:CQR_annotation}).
    }
    \label{table:ablation_label}
    \vspace{-6mm}
\end{table}

To investigate the impact of each component on the performance of \textsc{AdaCQR}, we conducted ablation experiments focusing on the below modules in Table~\ref{table:ablation} and Table~\ref{table:ablation_label}.

\paragraph{\colorbox{mycell}{Alignment}} 
The training Stage of \textsc{AdaCQR} incorporates a contrastive loss to align the retrievers.
To assess the influence of contrastive loss, we executed a single-stage training process without alignment.
The removal of contrastive loss results in the most notable decline in performance,  with a 9.1\% decrease in MRR metric.

\paragraph{\colorbox{mycelltwo}{Ranking}} 
We introduce a fusion metric to evaluate query performance across semantic and term perspectives. 
To determine the effect of the fusion metric, we substitute it with the metric used in previous work~\cite{IterCQR}, which ranks candidates solely based on the \textit{cosine similarity} between the candidate query and the gold passage.
The performance degradation confirms the effectiveness of using signals from the ranking of both types of retrievers.
To further investigate the effectiveness of considering both perspectives in the fusion metric, we separately remove sparse ranking $r_s$ and dense ranking $r_d$ within it for analysis.
Removing any rank degrades performance for both retrievers, more significantly for the corresponding retriever.
This confirms the rationale behind considering both perspectives simultaneously.

\paragraph{Reformulation Labels}
The performance of directly using human label and $Q^\star$ as the queries is reported in Table~\ref{table:ablation_label}.
It is worth noting that superior labels $Q^\star$ outperform human labels and have strong performance both in sparse and dense retrievals, which validates the effectiveness of the proposed fusion metric and the annotation method.
In the Training Stage 1 of \textsc{AdaCQR}, we use $Q^\star$ as the golden labels.
We also experiment by replacing it with human labels, and the resulting performance drop further validates the effectiveness of $Q^\star$.
Additionally, the final results of \textsc{AdaCQR} after both Training Stage 1 and Stage 2 show significant improvement over the original queries (whether $Q^\star$ or human labels), demonstrating the effectiveness of the Stage 2 alignment.

\subsection{Performance against CDR Methods}
\label{app:cdr}
Conversational Dense Retrieval (CDR) is an orthogonal method for conversational query reformulation in conversational search, which trains dense retrievers to improve the representation of the current query and historical context.
Although not directly comparable, we still present a performance comparison of the \textsc{AdaCQR} and CDR methods on the QReCC, TopiOCQA, and CAsT datasets, as shown in Table~\ref{table:cdr_comparsion}.

We compare our approach against four baseline models: Conv-ANCE~\cite{ANCE}, InstructoR-ANCE~\cite{instructor}, Conv-SPLADE~\cite{splade}, and LeCoRE~\cite{dense-retrieval-7}. 
Conv-ANCE trains a dense retriever by enhancing session representation using a conventional contrastive ranking loss. 
InstructoR-ANCE leverages large language models (LLMs) to predict the relevance score between sessions and passages, followed by retriever training. 
Conv-SPLADE fine-tunes a strong lexical-based retriever on conversational search data, utilizing ranking loss. 
LeCoRE extends the SPLADE model with multi-level denoising techniques to enhance lexical session representation.
\textsc{AdaCQR} achieves the best average performance across four datasets, demonstrating superiority in both effectiveness and generalizability.
Our approach employs the \textit{off-the-shelf}  ANCE retriever, and our rewrite method is \textbf{orthogonal} to the aforementioned retrievers, leaving the exploration of combining these methods for achieving higher performance in future research.
Moreover, while CDR methods focus on dense representations, there are scenarios where the sparse representation of BM25 demonstrates a retrieval advantage. 
Our approach \textsc{AdaCQR} takes both types of retrievers into account and enhances performance through a rewrite strategy.

\begin{table*}[t!]
    \centering
    \small
    \begin{threeparttable}
    \begin{tabular}{lccccc}
        \toprule
        \textbf{Method} & \textbf{TopiOCQA} & \textbf{QReCC} & \textbf{CAsT-20} & \textbf{CAsT-21} & \textbf{Avg.}\\
        \midrule
        Conv-ANCE~\cite{ANCE} & 22.9 & 47.1 & 42.2 & 52.3 & 41.1 \\
        InstructoR-ANCE~\cite{instructor} & 25.3 & 43.5 & 43.7 & 53.0 & 41.4 \\
        Conv-SPLADE~\cite{splade} & 30.7 & 50.0 & 36.9 & 47.9 & 41.4 \\
        LeCoRE~\cite{dense-retrieval-7} & 32.0 & 51.1 & 37.7 & 50.8 & 42.9 \\
        \textsc{AdaCQR}(\textit{Ours}) & 32.8 & 45.1 & 51.4 & 58.3 & \underline{46.9} \\
        \textsc{AdaCQR}\texttt{+Expansion}(\textit{Ours}) & 38.5 & 45.8 & 56.6 & 64.2 & \textbf{51.3} \\
        \midrule
        \bottomrule 
    \end{tabular}
    \end{threeparttable}
    \caption{
    MRR performance comparison of the \textsc{AdaCQR} and CDR methods.
    }
    \label{table:cdr_comparsion}
\end{table*}

\section{Analysis}
\subsection{Analysis of the Aligned Query}
\begin{figure}[t]
    \centering
    \includegraphics[width=0.5\textwidth]{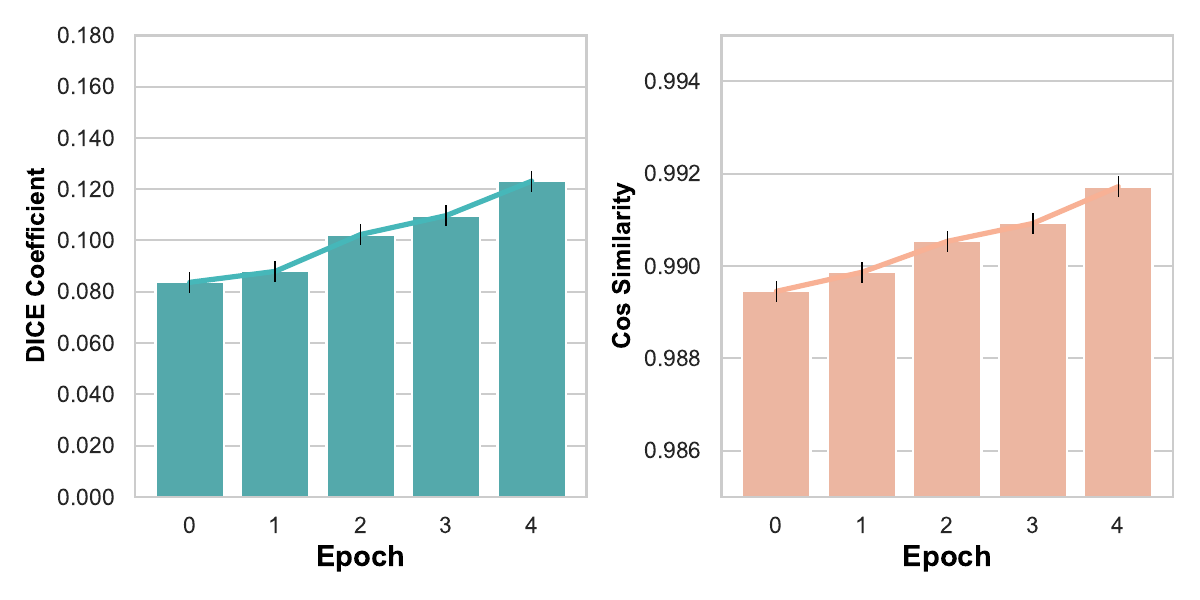}
    \caption{Analysis of the aligned reformulation query across epochs in Stage 2 training, focusing on the term overlap (DICE coefficient) and semantic similarity with the gold passage (cosine similarity).} 
    \label{fig:aligment_analysis}
    \vspace{-4mm}
\end{figure}
To evaluate the effectiveness of the aligned reformulation queries, we analyse the reformulation queries across the first 5 epochs during Stage 2 training in Figure~\ref{fig:aligment_analysis}. 
We conduct analyses focusing on the average term overlap and semantic similarity between the queries and the gold passages.
The DICE Coefficient~\cite{DICE_coefficient} is utilized to assess term overlap, while cosine similarity is employed to measure semantic similarity. 
This analysis indicates that both term overlap and semantic similarity between the reformulated queries and the gold passages exhibit an increasing trend with each epoch in Stage 2, demonstrating the effectiveness of our method in considering both perspectives.

\section{Conclusion}
In this paper, to achieve alignment between the reformulation model and both term and semantic retrieval systems, \textsc{AdaCQR} is proposed to enhance the generalizability of information-seeking queries through a two-stage training strategy.
By leveraging a fusion metric that assesses the generalization performance across different retrieval systems, we can effectively obtain superior labels for the generation and gather a diverse set of ordered candidate queries for reference-free evaluation.
Extensive experiments on five datasets demonstrate the superiority of \textsc{AdaCQR}, achieving performance comparable to methods using the fine-tuned LLaMA2-7B and proprietary LLM.

\section*{Limitations}
Although \textsc{AdaCQR} demonstrates remarkable performance in experimental evaluations, it also has several limitations.

During the \textsc{AdaCQR} training process, we leverage ChatGPT for superior reformulation label annotation, and our annotation prompt requires training a basic model, which incurs additional costs and training expenses. 
Furthermore, due to budget constraints, we do not use more powerful LLMs, such as GPT-4, to obtain reformulation labels, although it is obvious that employing a more powerful LLM would yield better reformulation labels.

Although no further costs are introduced during reformulation model inference, aligning AdaCQR with retrievers introduces additional training time.
Furthermore, generating the ordered candidate set for alignment demands extra retrieval time and increased storage capacity.

\section*{Acknowledgments}
The authors would like to thank the anonymous reviewers for their insightful comments. This work is funded by the National Natural Science Foundation of China (Grant No.62176053). This work is supported by the Big Data Computing Center of Southeast University.


\bibliography{references}
\clearpage
\newpage
\appendix

\section{Discussion}

\subsection{Advantage of \textsc{AdaCQR} in Alignment}\label{appendix:comparsion_alignment_methods}
Previous works like CONQRR~\cite{ConqRR} and IterCQR~\cite{IterCQR} explored the concept of alignment between the reformulation model and the retriever.
CONQRR leverages reinforcement learning to maximize the reward of the sampled rewrite $q_s$ and the baseline rewrite $q$.
IterCQR uses Minimum Bayes Risk (MBR) and Top-1 Selection to achieve the alignment between the reformulation model and dense retriever.

Compared with previous works, the main advantage of \textsc{AdaCQR} is that it \textbf{no longer requires an explicit reward model during training}; the language model serves simultaneously as both the generation model and the evaluation model (the evaluation model of CONQRR is BM25 retrieval system, and the evaluation model of IterCQR is a dense passage encoder). 
By decoupling the model training and candidates ranking (\ie, the candidates ranking is an offline process), our method achieves improved training efficiency.

Apart from this, \textsc{AdaCQR} achieves alignment based on contrastive learning, which is \textbf{easier to implement, converges faster, and is more stable during training} compared to the reinforcement learning(RL) of CONQRR and the Minimum Bayes Risk(MBR) of IterCQR ~\cite{BRIO}. 
Both CONQRR and IterCQR employ special techniques like min-max normalization to ensure model convergence, which increases the training complexity~\citep{ConqRR, IterCQR}.
Additionally, IterCQR requires two stages, MBR and Top-1 Selection, to achieve alignment in 7-10 epochs~\cite{IterCQR}, whereas \textsc{AdaCQR} achieves alignment in just 3-5 epochs.

\subsection{Effectiveness of Prompt Setting}
\begin{table}[h]
    \centering
    \small
    \begin{threeparttable}
    \begin{tabular}{clcc}
        \toprule
        & \multicolumn{3}{c}{\textbf{QReCC}}  \\ 
        \textbf{Type} & \textbf{Prompt Setting} & \textbf{MRR} & \textbf{R@10}\\
        \midrule
        \midrule 
        \multirow{3}{*}{\rotatebox[origin=c]{90}{\textbf{Sparse}}} 
        & 0-shot  & 36.3 & 54.9\\
        & 3-shot (\textbf{Random}) & 39.1 & 58.0 \\
        & 3-shot (\textbf{Representative}) & \textbf{45.4} & \textbf{65.5} \\
        \midrule
        \multirow{3}{*}{\rotatebox[origin=c]{90}{\textbf{Dense}}}
        & 0-shot  & 34.5 & 52.6\\
        & 3-shot (\textbf{Random}) & 37.2 & 56.0 \\
        & 3-shot (\textbf{Representative}) & \textbf{40.1} & \textbf{60.2} \\
        \midrule
        \bottomrule 
    \end{tabular}
    \end{threeparttable}
    \caption{
    The annotation results generated by ChatGPT under different prompt settings on the QReCC test set. 
    \textbf{Random} denotes examples randomly chosen from the validation set, while \textbf{Representative} refers to select examples as described in Section~\ref{sec:CQR_annotation}.
    }
    \vspace{-2mm}
    \label{table:ablation_prompt}
\end{table}

To evaluate the effectiveness of the prompt design method proposed in Section~\ref{sec:CQR_annotation}, we applied our prompt design method for reformulation label annotation on the QReCC test set.

We compared the results with the 0-shot approach (\ie, using only the Instruction and Annotated Sample parts from Table~\ref{prompt_for_qrecc}) and the 3-shot-random approach (\ie, randomly selecting 3 examples from the validation set). The results are shown in Table~\ref{table:ablation_prompt}.

Based on these results, our prompt setting significantly improves performance in both sparse and dense retrieval compared to the 0-shot and 3-shot-random methods, showing the effectiveness of our prompt setting.

\subsection{Effect of the Multi-Task Loss}
\begin{table}[t!]
    \centering
    \small
    \begin{threeparttable}
    \begin{tabular}{lcccc}
        \toprule
        & \multicolumn{4}{c}{\textbf{QReCC}} \\ 
        \textbf{Coefficient($\gamma$)} & \textbf{MRR} & \textbf{NDCG} & \textbf{R@10} & \textbf{R@100}\\
        \midrule
        0 & 43.3 & 41.0 & 62.8 & 88.5 \\
        0.1 & 45.3 & 42.7 & 65.2 & 90.2 \\
        1 & 48.8 & 46.1 & 68.7 & \textbf{91.2} \\
        10 & 50.2 & 47.7 & 68.8 & 89.0 \\
        100 & \textbf{52.4} & \textbf{49.9} & \textbf{70.9} & 91.0 \\
        1000 & 49.4 & 46.7 & 68.6 & 90.7 \\
        +$\infty$ & 44.5 & 41.8 & 65.5 & 90.9\\
        \midrule
        \bottomrule 
    \end{tabular}
    \end{threeparttable}
    \caption{
    \textsc{AdaCQR} performance with different $\gamma$ coefficients weighting of the contrastive loss in Eq.~\eqref{eq:stage2-loss}.
    +$\infty$ indicates only using the contrastive loss.
    0 indicates only using the cross-entropy loss. 
    BM25 is used as the retriever for experiments.
    }
    \label{table:gamma_coefficients_qrecc}
\end{table}

The multi-task loss defined in Eq.~\eqref{eq:stage2-loss} is designed to align with retrievers by incorporating both cross-entropy loss and contrastive loss. 
We conducted experiments with various $\gamma$ coefficients, as shown in Table~\ref{table:gamma_coefficients_qrecc}. 
The results indicate that increasing $\gamma$ improves the performance of \textsc{AdaCQR} within a certain range, highlighting the crucial role of contrastive loss for alignment. 
However, the importance of cross-entropy loss is also evident: when $\gamma$ is excessively high or cross-entropy loss is omitted, the performance declines. 
Therefore, it concludes that including cross-entropy loss is essential to prevent excessive model variation, illustrating its necessity in the design of this multi-task loss.

\subsection{Robustness to Topic Shifts in Conversation}
\begin{table}[t!]
    \centering
    \small
    \begin{threeparttable}
    \resizebox{\columnwidth}{!}{%
    \begin{tabular*}{1.03\columnwidth}{lcccc}
        \toprule
        & \multicolumn{2}{c}{\textbf{Topic-Concentrated}} & \multicolumn{2}{c}{\textbf{Topic-Shifted}}\\ 
        \textbf{Model} & \textbf{MRR} & \textbf{R@10} & \textbf{MRR} & \textbf{R@10}\\
        \midrule
        \midrule 
        T5QR & 35.2 & 54.4 & 25.2 & 45.1\\
        CONQRR & 41.9 & 63.1 & 25.2 & 45.9\\
        IterCQR & \underline{54.4} & \underline{72.4} & 24.9 & 49.7\\
        Human Rewrite & 44.0 & 66.7 & \underline{31.8} & \underline{56.7}\\
        \midrule
        \textsc{AdaCQR} & \textbf{66.0} & \textbf{82.4} & \textbf{34.1} & \textbf{58.3} \\
        \midrule
        \bottomrule 
    \end{tabular*}%
    }
    \end{threeparttable}
    \caption{
    Performance of \textsc{AdaCQR} on topic-concentrated and topic-shifted samples on QReCC, MRR and R@10 are reported.
    The result is reported on the BM25 Retrieval System.
    }
    \vspace{-2mm}
    \label{table:topic-shifts-qrecc}
\end{table}

In the conversational search task, the frequent topic changes during the dialogue pose challenges for CQR. 
To evaluate the robustness of \textsc{AdaCQR} in handling topic shifts, we divided the QReCC dataset into two parts: Topic-Concentrated and Topic-Shifted. 
Following previous work~\cite{IterCQR}, we determine whether a topic shift has occurred in the current conversation by checking if the gold passage ID associated with the current query appears in the gold passage IDs corresponding to the previous context.
The results presented in Table \ref{table:topic-shifts-qrecc} indicate that \textsc{AdaCQR} substantially outperforms previous models in both parts of conversations. 
Additionally, \textsc{AdaCQR} exceeds human rewrites in topic-shifted dialogues, showing the robustness of our approach in query reformulation when addressing topic shiftings.



\section{Experimental Details}
\subsection{Baseline Details}\label{appendix:baselines}
In our study, we compare \textsc{AdaCQR} with the following representative baselines in the CQR task:
\begin{itemize}[itemsep=0.5pt, parsep=0pt, left=-0.1em]
\item \textbf{T5QR}~\cite{T5QR} is a vanilla baseline that utilizes the T5-base~\cite{T5-paper} model to perform CQR tasks.

\item \textbf{CONQRR}~\cite{ConqRR} aligns the T5-base reformulation model with retrievers through direct optimization using reinforcement learning.

\item \textbf{ConvGQR} ~\cite{ConvGQR} enhances retrieval performance by employing two fine-tuned T5-base models, one for query reformulation and the other for query expansion.

\item \textbf{EDIRCS} ~\cite{EDIRCS} leverages the incorporation of non-autoregressive text-selecting techniques and autoregressive tokens generation to generate reformulation queries effectively on a fine-tuned T5-base model.

\item \textbf{LLM-Aided} ~\cite{InfoCQR} employs ChatGPT ~\cite{openaigpt35} to conduct query reformulation via a ``rewrite-then-edit'' prompting strategy.

\item \textbf{IterCQR} ~\cite{IterCQR} achieves the alignment of the T5-base reformulation model and dense retriever by minimizing Bayes Risk based on the semantic similarity between the query and the gold passage.

\item \textbf{\textsc{RetPO}}~\cite{RetPo} utilizes large language models to generate multiple reformulations through multi-perspective prompting, creates binarized comparisons based on retriever feedback, and optimizes LLaMA2-7B~\cite{llama2-paper} using direct preference optimization (DPO)~\cite{DPO}.

\item \textbf{LLM4CS}~\cite{LLM4CS} is the state-of-the-art conversational query reformulation baseline, utilizing LLM prompting techniques to generate multiple reformulated queries and aggregate their embeddings.
We adopt rewriting prompting (REW) and rewriting-and-response (RAR) as baselines for query reformulation and query reformulation with expansion, respectively.  
Our implementation of REW and RAR follows standard settings (\ie, generating five queries and aggregating them) but excludes chain-of-thought (CoT) content due to the extensive human annotation required, which is not part of our baselines and impractical in real-world scenarios.
It is important to note that aggregation techniques are applicable only in dense retrieval; therefore, for BM25 retrieval, we implement RAR by concatenating the query and response with the top-1 generation result.
\end{itemize}

\subsection{Datasets Details}\label{appendix:datasets}
\begin{table}[t]
    \centering
    \small
    \resizebox{\columnwidth}{!}{%
    \begin{tabular}{lcccccc}
    \toprule
    & \multicolumn{3}{c}{\textbf{QReCC}} &  \multicolumn{3}{c}{\textbf{TopiOCQA}}  \\ 
    & Train & Valid & Test & Train & Valid & Test \\
    \midrule
    \#\ Dialogues & 10822 & 769 & 2775 & 3509 & 720 & 205 \\
    \midrule
    \#\ Turns & 62701 & 800 & 16451 & 44650 & 800 & 2514\\
    \#\ Turns with Gold & 28796 & 800 & 8209 & 44650 & 800 & 2514\\
    \bottomrule
    \end{tabular}
    }
    \caption{The statistics of QReCC and TopiOCQA datasets. }
    \label{table:dataset_details}
\end{table}

The QReCC~\cite{qrecc-datasets} dataset comprises 14K conversations with 80K question-answer pairs, and we aim to retrieve the gold passage from a collection containing 54M passages. 
Conversely, the TopiOCQA~\cite{topiocqa-datasets} dataset includes 3.9K topic-switching conversations with 51K question-answer pairs, where the passage collection is sourced from Wikipedia and contains about 20M passages. 
Notably, a few examples from the QReCC and TopiOCQA training sets were randomly partitioned to create respective validation sets.
The details of QReCC and TopiOCQA are described in Table \ref{table:dataset_details}.

TREC CAsT 2019, 2020, and 2021~\cite{cast19,cast20, cast21} are datasets known for their complexity and challenges in conversational search with a zero-shot setting. 
The statistics of CAsT 19-21 are shown in Table~\ref{table:cast_dataset_details}.

\begin{table}[h]
    \centering
    \small
    \resizebox{\columnwidth}{!}{%
    \begin{tabular}{lccc}
    \toprule
    & CAsT-19 &  CAsT-20 & CAsT-21  \\ 
    \midrule
    \# Dialogues & 50 & 25 & 26 \\
    \# Turns & 479 & 208 & 239 \\
    \midrule
    \# Collections & 38M & 38M & 40M \\
    \bottomrule
    \end{tabular}
    }
    \caption{The statistics of CAsT 2019, 2020, and 2021 datasets. }
    \label{table:cast_dataset_details}
\end{table}

\subsection{Evaluation Metrics}
\label{appendix:eval}
We evaluate \textsc{AdaCQR}'s retrieval performance using several widely used metrics, such as Mean Reciprocal Rank (MRR), Normalized Discounted Cumulative Gain (NDCG), Recall@10, and Recall@100. 
MRR is a ranking quality metric that considers the position of the first relevant passage among the ranked passages. 
NDCG@3 evaluates the retrieval results by considering the relevance and the rank of the top three results. 
Recall@K measures whether the gold passage is present within the top-K results.

\section{Implementation Details}\label{appendix:implementation_details}
All experiments are conducted on a server equipped with four Nvidia GeForce 3090 GPUs.
\subsection{\textsc{AdaCQR} Details}
For the implementatio of \textsc{AdaCQR}, we use Huggingface \textit{transformers} library \footnote{\url{https://github.com/huggingface/transformers}} and \textit{Pytorch Lightning} \footnote{\url{https://github.com/Lightning-AI/pytorch-lightning}} framework.

We use T5-base\footnote{\url{https://huggingface.co/google-t5/t5-base}}~\cite{T5-paper} as the backbone of \textsc{AdaCQR}. 
After conducting a comprehensive grid search, we configured the number of candidates $n=32$, the margin parameter $\lambda=0.1$, the weight of the contrastive loss $\gamma=100$, the length penalty parameter $\alpha=0.6$, and the probability mass parameter in label smooth distribution $\beta = 0.1$. 
The model parameters are optimized by the \texttt{AdamW} optimizer~\cite{adamw}. 

\textsc{AdaCQR} is trained for $10$ epochs in Stage 1 with a learning rate set to \texttt{2e-5} and 8 epochs in Stage 2 with a learning rate adjusted to \texttt{5e-6}.
Both stages incorporate linear learning rate schedulers with a warm-up ratio of $0.1$. 
We employ grid search for hyperparameter selection. 

The vanilla reformulation model $G_\pi$ in Section~\ref{sec:CQR_annotation} is trained on reformulation labels of the QReCC dataset acquired by zero-shot prompting with ChatGPT, and the prompt is shown in Appendix~\ref{appendix:annotation}.
This model is trained in 10 epochs, and the learning rate is set to \texttt{2e-5} with a linear learning rate scheduler with a warm-up ratio of 0.1.

For candidate generation in Section~\ref{sec:candidates_generation}, we used diverse beam search with a diverse penalty of 2.0. The minimum token length for generated candidates is set to 8, and the maximum token length is set to 64. For the generation of reformulation queries, we employed beam search with a beam size of 5, and the maximum token length is set to 64 for generated queries.

\subsection{Retrieval Systems Details}
We implement the retrieval systems using Faiss~\cite{faiss} and Pyserini~\cite{pyserini}. 
For BM25, as in previous work~\cite{ConvGQR, IterCQR, RetPo}, we set $k_1=0.82$, $b = 0.68$ in QReCC, and $k_1=0.9$, $b=0.4$ in TopiOCQA. 
The $k_1$ controls the non-linear term frequency normalization, and $b$ is the scale of the inverse document frequency.
For ANCE\footnote{\url{https://huggingface.co/sentence-transformers/msmarco-roberta-base-ance-firstp}}, the maximum token length is set to 128 tokens for reformulation query and 384 tokens for passage.

For both sparse and dense retrieval systems, we retrieved the top 100 relevant passages for each query and obtained the result of evaluation metrics with \textit{pytrec\_eval}~\cite{pytrec_eval}.

\section{ChatGPT Annotation Details}
\label{appendix:annotation}
We use \texttt{gpt-3.5-turbo-0125}~\citep{openaigpt35}\footnote{\url{https://platform.openai.com/docs/models/gpt-3-5-turbo}} to obtain the initial and superior reformulation labels via zero-shot and few-shots prompting.

For initial reformulation labels of $G_\pi$, we use the ``Instruction'' and ``Annotated Sample'' parts shown in Table~\ref{prompt_for_qrecc}, \ie, zero-shot.

For superior reformulation labels for \textsc{AdaCQR}, we utilize the top-3 most challenging demonstrations (\ie, $m=3$) for the QReCC dataset and the top-5 most challenging demonstrations (\ie, $m=5$) for the TopiOCQA dataset, \ie, few-shots.
The final prompt is formed $\mathcal{D} = \mathcal{I}\ || T_1 || \cdots || T_m$, where $||$ donates concatenation. 
The detailed prompts to annotate the QReCC dataset and the TopiOCQA dataset are shown in Table~\ref{prompt_for_qrecc} and Table~\ref{prompt_for_topiocqa}, respectively.

To encourage a more deterministic output, we set the temperature to $0.1$, and the seed is set to $42$ for reproductivity.
The total consumption to annotate QReCC and TopiOCQA datasets for initial and superior reformulation labels is about 151M tokens, which cost about 120\$.

\section{Case Study}\label{appendix:case_study}
In this section, we present several examples of how \textsc{AdaCQR} succeeded or failed on the QReCC and TopiOCQA datasets. 

Table~\ref{table:case_study_qrecc_bm25_success} demonstrates a case where \textsc{AdaCQR} successfully retrieved the gold passage through query rewriting, whereas human rewrites failed, showing the superiority of \textsc{AdaCQR} over human rewrites. 
After being written by \textsc{AdaCQR}, the query is decontextualized, resulting in overlaps while concurrently offering more specific information.
This enhanced specificity aids the retriever toward the most relevant passages effectively.
Additionally, in Tables~\ref{table:case_study_topiocqa_bm25_success} and~\ref{table:case_study_topiocqa_bm25_expansion}, we show examples of how the \textsc{AdaCQR} and \textsc{AdaCQR} with Expansion models successfully retrieved the gold passage.
We also provide a failure case, as illustrated in Table~\ref{table:case_study_qrecc_bm25_fail}.

\section{Query Expansion Details}\label{appendix:query_expansion}
For query expansion, we leverage \texttt{LLaMA2-7B-Chat}\footnote{\url{https://huggingface.co/meta-llama/Llama-2-7b-chat-hf}} as the backbone for a fair comparison with prior work~\cite{RetPo}. 
The query expansion process involves directly answering the given query~\cite{ConvGQR} and generating relevant keywords~\cite{keywords_query_expansion}. 
Then the reformulation queries are concated with the generated answers and keywords for retrieval.
The prompts employed for query expansion are presented in Table~\ref{expansion_prompt_answer} and Table~\ref{expansion_prompt_keywords}.

\textit{vLLM} framework~\cite{VLLM} is used for inference, with the temperature parameter set to 0.5 and the maximum token limit set to 50 during the generation process.

\begin{table*}
\begin{tcolorbox}[
pad at break*=1mm,
colback=white!95!gray,
colframe=gray!50!black,
title=Prompt for QReCC Annotation]
\textbf{Instruction}
\begin{lstlisting}[breaklines=true, xleftmargin=0pt, breakindent=0pt, columns=fullflexible]
Given a question and its context, decontextualize the question by addressing coreference and omission issues. The resulting question should retain its original meaning and be as informative as possible, and should not duplicate any previously asked questions in the context.
\end{lstlisting}

\textbf{Demonstrations}
\begin{lstlisting}[breaklines=true, xleftmargin=0pt, breakindent=0pt, columns=fullflexible]
Context: [Q: What was Ridley Scott's directing approach to directing? A: Russell Crowe commented about Ridley Scott's directing, I like being on Ridley's set because actors can perform and the focus is on the performers. Q: Were there others who commented about Scott's approach as a director and producer? A: Charlize Theron praised the Ridley Scott's willingness to listen to suggestions from the cast for improvements in the way their characters are portrayed on screen. Q: What was Ridley Scott's style? A: In Ridley Scott's visual style, he incorporates a detailed approach to production design and innovative, atmospheric lighting Q: How did that translate into his films? A: In his movies, Ridley Scott commonly uses slow pacing until the action sequences. Q: What popular movies did he take this approach and use this style? A: Examples of Ridley Scott's directing style include Alien and Blade Runner.]
Question: Is there anything else interesting about his style?
Good Rewrite: Is there anything else interesting about Ridley Scott's style besides his slow pacing until the action sequences?
Bad Rewrite: is there anything else interesting about Ridley Scott's directing style?

Context: [Q: What was the health issues did Bad Brains frontman H.R. have? A: On March 15, 2016, Bad Brains frontman H.R. was reportedly diagnosed with a rare type of headache called Short-lasting unilateral neuralgiform headache with conjunctival injection and tearing (SUNCT syndrome) Q: Was there anything to cure it? A: As diagnostic criteria have been indecisive and its pathophysiology remains unclear, no permanent cure is available for short-lasting unilateral neuralgiform headache with conjunctival injection and tearing (SUNCT syndrome) Q: Are there any other interesting aspects about this article? A: On November 3, 2015, Bad Brains announced on their Facebook page that Dr. Know (Gary Miller) was hospitalized and on life support, after many other musicians reported so.]
Question: What did they do in 2015?
Good Rewrite: What did Bad Brains do in 2015 after Dr. Know (Gary Miller) was hospitalized and on life support?
Bad Rewrite: What do the Bad Brains do in 2015?

(*@\textcolor{gray}{<--- Omit One demonstration --->}@*)
\end{lstlisting}

\textbf{Annotated Sample}
\begin{lstlisting}[breaklines=true, xleftmargin=0pt, breakindent=0pt, columns=fullflexible]
Context: [(*@\textcolor{c1}{
\{\{current\_context\}\}
}@*)]
Question: (*@ \textcolor{c1}{
\{\{current\_query\}\}
}@*)
Good Rewrite:
\end{lstlisting}
\end{tcolorbox}
\caption{The prompt used to obtain QReCC annotated labels.}
\label{prompt_for_qrecc}
\end{table*}

\begin{table*}
\begin{tcolorbox}[pad at break*=1mm,colback=white!95!gray,colframe=gray!50!black,title=Prompt for TopiOCQA Annotation]
\textbf{Instruction}
\begin{lstlisting}[breaklines=true, xleftmargin=0pt, breakindent=0pt, columns=fullflexible]
Given a question and its context, decontextualize the question by addressing coreference and omission issues. The resulting question should retain its original meaning and be as informative as possible, and should not duplicate any previously asked questions in the context.
\end{lstlisting}

\textbf{Demonstrations}
\begin{lstlisting}[breaklines=true, xleftmargin=0pt, breakindent=0pt, columns=fullflexible]
Context: [Q: what is the fallacy of the argumentum ad hominem A: That it is not always fallacious, and that in some instances, questions of personal conduct, character, motives, etc., are legitimate and relevant to the issue, as when it directly involves hypocrisy, or actions contradicting the subject's words. Q: what does that last phrase mentioned above mean? A: It is an argumentum(a quarrel; altercation) ad hominem, refers to several types of arguments, not all are fallacious. Q: where does this phrase come from? A: The ancient Greek. Q: are there any philosophers who have written about this? A: Yes, Greeks. Aristotle, Sextus Empiricus, John Locke, Charles Leonard Hamblin, Douglas N. Walton. Q: who is the first mentioned person? A: He was a Greek philosopher and polymath during the Classical period in Ancient Greece. Q: has he written any book? A: He has written on subjects including physics, biology, zoology, metaphysics, logic, ethics, aesthetics, poetry, theatre, music, rhetoric, psychology, linguistics, economics, politics, and government. Q: what did he theorize about dreaming? A: He explained that dreams do not involve actually sensing a stimulus. In dreams, sensation is still involved, but in an altered manner.He also explains that when a person stares at a moving stimulus such as the waves in a body of water, and then look away, the next thing they look at appears to have a wavelike motion. Q: who is the second philosopher mentioned earlier? A: Sextus Empiricus was a Pyrrhonist philosopher and a physician mostly involved in ancient Greek and Roman Pyrrhonism.]
Question: do his teachings/work have any similarities with buddhism?
Good Rewrite: do sextus empiricus' teachings/work have any similarities with buddhism?
Bad Rewrite: there are any similarities between the philosophers mentioned above and buddhism.

Context: [Q: who was the french leader the diplomats were trying to meet with A: French foreign minister Talleyrand Q: what was this affair about? A: Confrontation between the United States and Republican France that led to the Quasi-War. Q: what was this confrontation about? A: To negotiate a solution to problems that were threatening to break out into war. Q: can you name any one who attended the previous meetings? A: Charles Cotesworth Pinckney Q: who was he? A: He was an early American statesman of South Carolina, Revolutionary War veteran, and delegate to the Constitutional Convention. Q: where was he born? A: Charleston, South Carolina]
Question: what was his views regarding slaves?
Good Rewrite: what was charles cotesworth pinckney's views regarding slaves?
Bad Rewrite: whatever were carlos castellanos' views regarding slaves?

(*@\textcolor{gray}{<--- Omit Three Demonstrations --->}@*)
\end{lstlisting}

\textbf{Annotated Sample}
\begin{lstlisting}[breaklines=true, xleftmargin=0pt, breakindent=0pt, columns=fullflexible]
Context: [(*@\textcolor{c1}{
\{\{current\_context\}\}
}@*)]
Question: (*@ \textcolor{c1}{
\{\{current\_query\}\}
}@*)
Good Rewrite:
\end{lstlisting}
\end{tcolorbox}
\caption{The prompt used to obtain TopiOCQA annotated labels.}
\label{prompt_for_topiocqa}
\end{table*}

\begin{table*}
\begin{tcolorbox}[pad at break*=1mm,colback=white!95!gray,colframe=gray!50!black,title=Prompt for Query Expansion (Answer)]
\textbf{Instruction}
\begin{lstlisting}[breaklines=true, xleftmargin=0pt, breakindent=0pt, columns=fullflexible]
Given a question, please answer the question in a sentence. The answer should be as informative as possible.
\end{lstlisting}

\textbf{Demonstrations}
\begin{lstlisting}[breaklines=true, xleftmargin=0pt, breakindent=0pt, columns=fullflexible]
Question: and by whom was the game the last of us established?
Answer: Andy Gavin and Jason Rubin. Naughty Dog, LLC (formerly JAM Software, Inc.) is an American first-party video game developer based in Santa Monica, California. Founded by Andy Gavin and Jason Rubin in 1984 as an independent developer.

Question: is chelsea a club?
Answer: Yes, chelsea is an English professional football club. 

Question: is call me by your name a movie?
Answer: Yes, based on a book of the same name. Call Me by Your Name is a 2017 coming-of-age romantic drama film directed by Luca Guadagnino. Its screenplay, by James Ivory, who also co-produced, is based on the 2007 novel of the same name by Andr Aciman.

Question: where was alan menken born?
Answer: lan Irwin Menken was born on July 22, 1949, at French Hospital in Manhattan, to Judith and Norman Menken.

Question: where was ulysses s. grant from?
Answer: Hiram Ulysses Grant was born in Point Pleasant, Ohio, on April 27, 1822, to Jesse Root Grant, a tanner and merchant, and Hannah Simpson Grant.
\end{lstlisting}

\textbf{Annotated Sample}
\begin{lstlisting}[breaklines=true, xleftmargin=0pt, breakindent=0pt, columns=fullflexible]
Question: (*@ \textcolor{c1}{
\{\{reformulation\_query\}\}
}@*)
Answer:
\end{lstlisting}
\end{tcolorbox}
\caption{The prompt for query expansion by directly answering the question.}
\label{expansion_prompt_answer}
\end{table*}

\begin{table*}
\begin{tcolorbox}[pad at break*=1mm,colback=white!95!gray,colframe=gray!50!black,title=Prompt for Query Expansion (Keywords)]
\textbf{Instruction}
\begin{lstlisting}[breaklines=true, xleftmargin=0pt, breakindent=0pt, columns=fullflexible]
Write a few keywords for the given query.
\end{lstlisting}

\textbf{Annotated Sample}
\begin{lstlisting}[breaklines=true, xleftmargin=0pt, breakindent=0pt, columns=fullflexible]
Query: (*@ \textcolor{c1}{
\{\{reformulation\_query\}\}
}@*)
Keywords:
\end{lstlisting}
\end{tcolorbox}
\caption{The prompt for query expansion by giving keywords.}
\label{expansion_prompt_keywords}
\end{table*}

\begin{table*}
    \centering
    \resizebox{\textwidth}{!}{
    \def\arraystretch{0.9}
    \begin{tabularx}{\textwidth}{X}
        \toprule
            \textbf{Conversation}:\\
            Q1: What was the Securities Act of 1933? \\
            A1: The Securities Act of 1933 has two basic objectives: To require that investors receive financial and other significant information concerning securities being offered for public sale; and. To prohibit deceit, misrepresentations, and other fraud in the sale of securities. \\
            Q2: What is exempt from it? \\
            A2: However, there are exempt securities, under Section 4 of the Securities Act of 1933. These securities are financial instruments that carry government backing and typically have a government or tax-exempt status \\
            Q3: Why was it needed? \\
            A3: The act took power away from the states and put it into the hands of the federal government. The act also created a uniform set of rules to protect investors against fraud. \\
            Q4: What was the reason for creating the 1934 act? \\
            A4: The SEA of 1934 was enacted by Franklin D. Roosevelt's administration as a response to the widely held belief that irresponsible financial practices were one of the chief causes of the 1929 stock market crash.\\
            Q5: What is the largest securities exchange in the world? \\
            A5: The New York Stock Exchange founded on May 17, 1792, is the world's biggest stock exchange in trader value and has a capitalization of \$19.223 Trillion USD. \\
            \textbf{Original Query}: How and when when was it created? (\textbf{rank: Not Found}) \\\\
            \textbf{Human Rewrite}: How and when was the largest securities exchange in the world created? (\textbf{rank: Not Found})\\\\
            \textbf{\textsc{AdaCQR}}(\textit{Ours}): How and when was \underline{the Securities Act of 1933} created? The New York Stock Exchange founded on May 17, 1792? (\textbf{rank: 2})\\\\
        \bottomrule 
    \end{tabularx}}
    \caption{
    Case study on QReCC (id: 8\_6) when using BM25. The \underline{underline} part shows the decontextualized information in the reformulation query. We do not show the gold passage of this case because it is too long.
    }
    \label{table:case_study_qrecc_bm25_success}
\end{table*}

\begin{table*}[t!]
    \centering
    \resizebox{\textwidth}{!}{
    \def\arraystretch{0.9}
    \begin{tabularx}{\textwidth}{X}
        \toprule
            \textbf{Conversation}:\\
            Q1: when did the first episode of sesame street air? \\
            A1: 10 November 1969. \\
            Q2: is it a series? \\
            A2: Yes, an educational children's television series. \\
            Q3: is the series still running? \\
            A3: Yes, it is on its second run. \\
            Q4: when was the pilot episode done? \\
            A4: July 1969. \\
            Q5: is it a puppet show? \\
            A5: Yes. \\
            Q6: name a few characters from the series? \\
            A6: Human characters included Susan and muppet characters included Elmo. \\
            Q7: can you name more muppet characters? \\
            A7: Big Bird and Oscar the Grouch. \\
            Q8: how do the latter look like? \\
            A8: He has a green body with no visible nose. \\
            Q9: does the muppet perform any oscar related play? \\
            A9: UNANSWERABLE \\
            Q10: who performed the aforementioned bird muppet? \\
            A10: It was performed by Caroll Spinney till his retirement. \\
            Q11: who is he by profession? \\
            A11: He was an American puppeteer, cartoonist, author and speaker. \\
            \textbf{Original Query}: did he do comics too? (\textbf{rank: Not Found}) \\\\
            \textbf{\textsc{AdaCQR}}(\textit{Ours}): did \underline{Caroll Spinney} do \underline{Caroll} comics? (\textbf{rank: 1})\\\\
            \textbf{Gold Passage}: Caroll Spinney Comics and cartoons While in the Air Force, Spinney wrote and illustrated "Harvey", a comic strip about military life. He also animated a series of black-and-white cartoons called "Crazy Crayon".\\\\
        \bottomrule 
    \end{tabularx}}
    \caption{
    Successful case study on TopiOCQA (id: 16\_12) when using BM25. The \underline{underline} part shows the decontextualized information in the reformulation query.
    }
    \label{table:case_study_topiocqa_bm25_success}
\end{table*}

\begin{table*}[t!]
    \centering
    \resizebox{\textwidth}{!}{
    \def\arraystretch{0.9}
    \begin{tabularx}{\textwidth}{X}
        \toprule
        \textbf{Conversation}:\\
        Q1: does callie baby die in season 7 episode 18?\\
        A1: No. \\
        Q2: who plays the character mentioned above? \\
        A2: Sara Ramirez. \\
        Q3: apart from acting, does she have a career in any other profession? \\
        A3: She is a singer and songwriter. \\
        Q4: name some of her songs ? \\
        A4: Silent Night. \\
        Q5: what is the significance of the above song? \\
        A5: It is a popular Christmas carol. \\
        Q6: who has written it? \\
        A6: Joseph Mohr \\
        Q7: the above mentioned episode is from which series? \\
        A7: "Grey's Anatomy" \\
        Q8: name some characters of it. \\
        A8: Meredith Grey, Alex Karev, Miranda Bailey and Richard Webber \\
        Q9: what is the real name of the third character mentioned in the above list? \\
        A9: Chandra Wilson \\
        Q10: which movie did she debute in? \\
        A10: "Philadelphia" \\
        \textbf{Original Query}: what was it about? (\textbf{rank: Not Found}) \\\\
        \textbf{\textsc{AdaCQR}}: what was the movie "Philadelphia" about? (\textbf{rank: Not Found})\\\\
        \textbf{\textbf{AdaCQR} + Expansion}: what was the movie "Philadelphia" about? \textcolor{red}{Philadelphia is a 1993 American drama film directed by Jonathan Demme and starring Tom Hanks and Denzel Washington. The movie tells the story of Andrew Beckett, a gay lawyer who is fired from his job because of his sexual orientation, and his subsequent fight for justice and equality in the legal system.} \textcolor{blue}{Philadelphia, movie, Tom Hanks, Denzel Washington, AIDS, discrimination, lawsuit.} (\textbf{rank: 1})\\\\
        \textbf{Gold Passage}: Philadelphia (film) Introduction Philadelphia is a 1993 American legal drama film written by Ron Nyswaner, directed by Jonathan Demme and starring Tom Hanks and Denzel Washington. It was one of the first mainstream Hollywood films to acknowledge HIV/AIDS, homosexuality, and homophobia. For his role as Andrew Beckett, Hanks won the Academy Award for Best Actor at the 66th Academy Awards, while the song "Streets of Philadelphia" by Bruce Springsteen won the Academy Award for Best Original Song. Nyswaner was also nominated for the Academy Award for Best Original Screenplay, but lost to Jane Campion for "The Piano". \\\\
        \bottomrule
    \end{tabularx}}
    \caption{
    Successful case study with query expansion on TopiOCQA (id: 55\_11) when using BM25. The \textcolor{red}{part} and the \textcolor{blue}{part} represent the answers and keywords generated by LLM, respectively. These components furnish additional information that assists the retriever in enhancing its performance.
    }
    \label{table:case_study_topiocqa_bm25_expansion}
\end{table*}

\begin{table*}[t!]
    \centering
    \resizebox{\textwidth}{!}{
    \def\arraystretch{0.9}
    \begin{tabularx}{\textwidth}{X}
        \toprule
            \textbf{Conversation}:\\
            Q1: How can you tell if someone is suffering from depression?\\
            A1: You may be depressed if, for more than two weeks, you've felt sad, down or miserable most of the time, or have lost interest or pleasure in usual activities, and have also experienced several of the signs and symptoms across at least three of the categories below.\\
            Q2: What are common types? \\
            A2: he four most common types of depression are major depression, persistent depressive disorder(formerly known as dysthymia), bipolar disorder, and seasonal affective disorder. \\
            Q3: What causes it? \\
            A3: Rather, there are many possible causes of depression, including faulty mood regulation by the brain, genetic vulnerability, stressful life events, \\
            Q4: What is the role of brain chemicals? \\
            A4: A chemical imbalance in the brain is said to occur when there’s either too much or too little of certain chemicals, called neurotransmitters, in the brain. Neurotransmitters are natural chemicals that help facilitate communication between your nerve cells.\\
            Q5: Does a lack of sunlight cause it?\\
            A5: If you're not careful, a lack of sunlight can actually lead to a form of clinical depression.\\
            \textbf{Original Query}: How can you treat SAD? (\textbf{rank: 2}) \\\\
            \textbf{Human Rewrite}: How can you treat SAD? (\textbf{rank: 2})\\\\
            \textbf{\textsc{AdaCQR}}(\textit{Ours}): How can SAD be treated if a lack of sunlight in the brain can actually lead to depression? (\textbf{rank: Not Found})\\\\
        \bottomrule 
    \end{tabularx}}
    \caption{
    Failure case on QReCC (id: 57\_6) when using BM25.
    }
    \label{table:case_study_qrecc_bm25_fail}
\end{table*}

\end{document}